\documentclass[conference]{IEEEtran}
\IEEEoverridecommandlockouts
\usepackage{cite}
\usepackage{amsmath,amssymb,amsfonts}
\usepackage{algorithmic}
\usepackage{graphicx}
\usepackage{textcomp}
\usepackage{xcolor}
\def\BibTeX{{\rm B\kern-.05em{\sc i\kern-.025em b}\kern-.08em
    T\kern-.1667em\lower.7ex\hbox{E}\kern-.125emX}}

\makeatletter 
\newcommand{\linebreakand}{%
  \end{@IEEEauthorhalign}
  \hfill\mbox{}\par
  \mbox{}\hfill\begin{@IEEEauthorhalign}
}
\makeatother 
\begin{document}

\title{Privacy-Preserving Hybrid Ensemble Model for Network Anomaly Detection: Balancing Security and Data Protection\\
}

\author{
\IEEEauthorblockN{Shaobo Liu*}
\IEEEauthorblockA{\textit{Independent Researcher} \\
 Broomfield, USA \\
shaobo1992@gmail.com}
\and
\IEEEauthorblockN{Zihao Zhao}
\IEEEauthorblockA{\textit{Stevens Institute of Technology} \\
 Hoboken, USA \\
zzhao7189@gmail.com}
\and
\IEEEauthorblockN{Weijie He}
\IEEEauthorblockA{\textit{UCLA} \\
Los Angeles, USA \\
hahoou59@gmail.com}
\and
\linebreakand
\IEEEauthorblockN{Jiren Wang}
\IEEEauthorblockA{\textit{Virginia Tech} \\
Virginia, USA \\
jiren@vt.edu}
\and
\IEEEauthorblockN{Jing Peng}
\IEEEauthorblockA{\textit{University of Southern California} \\
Los Angeles, USA \\
jingpeng@usc.edu}
\and
\IEEEauthorblockN{Haoyuan Ma}
\IEEEauthorblockA{\textit{Independent Researcher} \\
San Jose, USA \\
hmabupt@gmail.com}
}

\maketitle

\begin{abstract}
Privacy-preserving network anomaly detection has become an essential area of research due to growing concerns over the protection of sensitive data. Traditional anomaly detection models often prioritize accuracy while neglecting the critical aspect of privacy. In this work, we propose a hybrid ensemble model that incorporates privacy-preserving techniques to address both detection accuracy and data protection. Our model combines the strengths of several machine learning algorithms, including K-Nearest Neighbors (KNN), Support Vector Machines (SVM), XGBoost, and Artificial Neural Networks (ANN), to create a robust system capable of identifying network anomalies while ensuring privacy. The proposed approach integrates advanced preprocessing techniques that enhance data quality and address the challenges of small sample sizes and imbalanced datasets. By embedding privacy measures into the model design, our solution offers a significant advancement over existing methods, ensuring both enhanced detection performance and strong privacy safeguards.

\end{abstract}


\section{Introduction}
The rise in sophisticated network attacks demands more effective anomaly detection systems that focus on deviations from normal behavior. Traditional methods struggle with evolving threats and large, imbalanced datasets where normal traffic vastly outnumbers malicious activity, leading to biased models. Additionally, the use of sensitive network data raises privacy concerns, requiring solutions that balance detection accuracy with data protection.

To address these challenges, we propose a privacy-preserving ensemble model combining K-nearest neighbors, support vector machines, XGBoost, and artificial neural networks. This approach leverages the strengths of each algorithm while employing synthetic data to enhance performance on imbalanced datasets and safeguard privacy. By integrating diverse methods and privacy-preserving techniques, our framework offers a robust and secure solution for anomaly detection in modern networks.

\section{Related Work}
Recent advancements in privacy-preserving anomaly detection have focused on federated learning (FL), ensemble models, and homomorphic encryption to secure data while maintaining high detection accuracy. Yang et al.\cite{yang2023achieving} developed a federated XGBoost model, effectively balancing privacy and performance in cross-silo anomaly detection by applying differential privacy in decentralized systems.

For addressing imbalanced datasets, Maniriho et al.\cite{maniriho2020anomaly} employed SMOTE to balance IoT datasets and improve machine learning model performance for detecting rare malicious events, while Liu et al.\cite{liu2020anomaly} demonstrated the effectiveness of hybrid SVM-KNN models for capturing diverse anomaly types.

However, federated learning models are vulnerable to poisoning attacks, as Nguyen et al.\cite{nguyen2020poisoning} highlighted, proposing strategies to defend FL-based systems in IoT networks from such threats. Shen et al.\cite{shen2024financial} explores using FinBERT and GPT-4o for financial sentiment analysis, showing that GPT-4o with few-shot learning rivals a fine-tuned FinBERT.

These studies collectively highlight innovations in federated learning, ensemble methods, and privacy-enhancing techniques for secure and effective anomaly detection across diverse networks.

\section{Data Preprocessing}  
Effective preprocessing is vital for machine learning, particularly in network intrusion detection, where datasets often suffer from noise, imbalance, and limited size. Key tasks include anomaly analysis, sample balancing, missing data management, and privacy safeguards.

\subsection{Anomaly Detection and Distribution Analysis}  
\begin{figure}[htbp]
\centering
\includegraphics[width=0.45\textwidth]{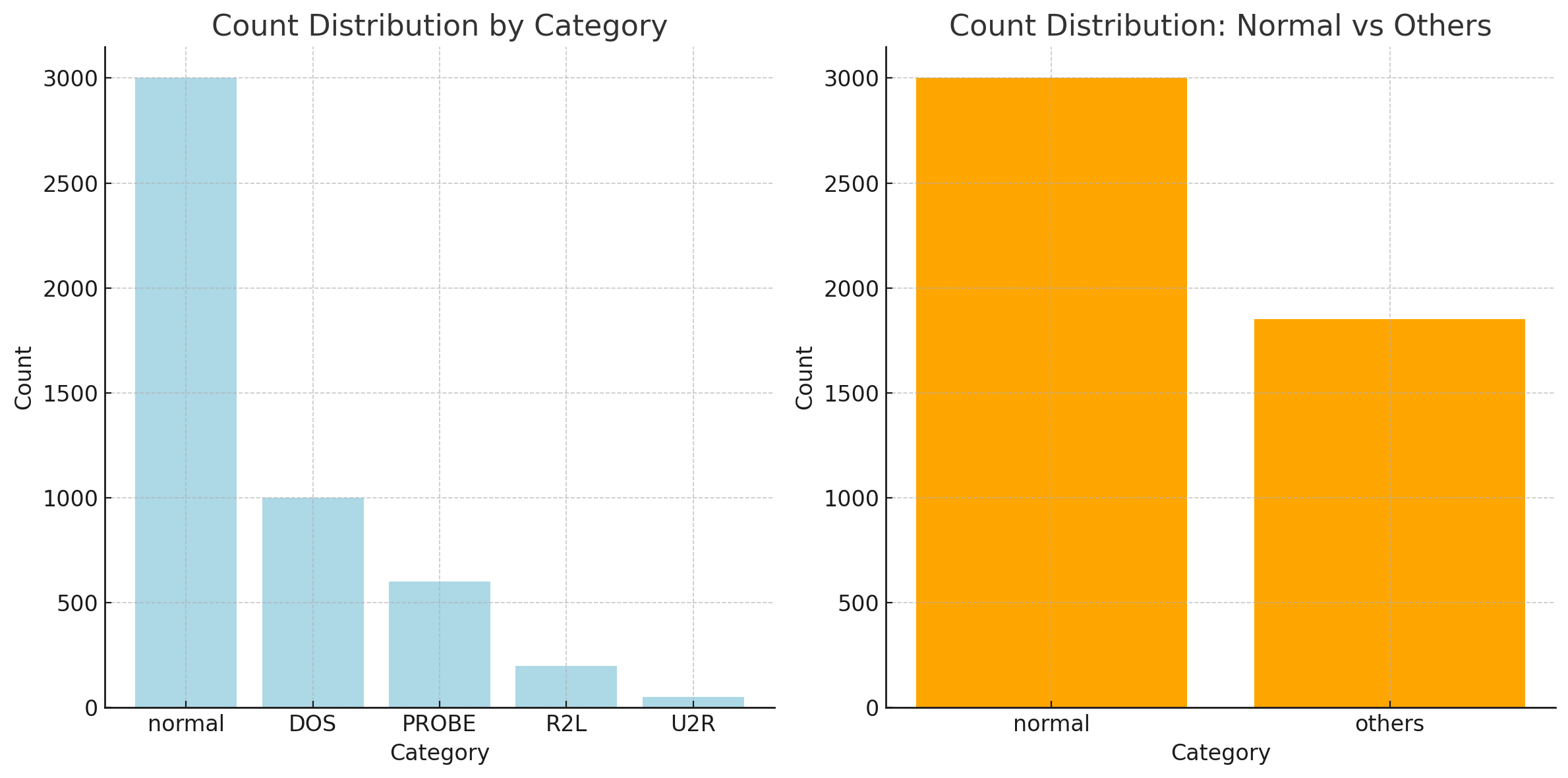}
\caption{Distribution of network anomalies.}
\label{fig:data-1}
\end{figure}

In preprocessing, analyzing anomaly distribution is critical. The dataset comprises normal traffic and attack types such as DoS, Probe, R2L, and U2R. The class proportion $P_i$ for each anomaly class $C_i$ is:
\begin{equation}
P_i = \frac{|C_i|}{|C|}
\end{equation}
where $|C_i|$ and $|C|$ denote instances in class $i$ and total instances. Figure \ref{fig:data-1} presents the distribution of anomalies, revealing a significant class imbalance, with normal and DoS dominating, while R2L and U2R are sparse, complicating unbiased training.

\subsection{Sample Balancing with Advanced Techniques}  
Due to the dataset's limited size and imbalance, traditional oversampling was insufficient. We applied advanced techniques, including small-sample learning and similarity-based sampling.

\textbf{Small-Sample Learning:}  
This technique enhanced generalization by generating synthetic samples with slight variations.

\begin{figure}[htbp]
\centering
\includegraphics[width=0.45\textwidth]{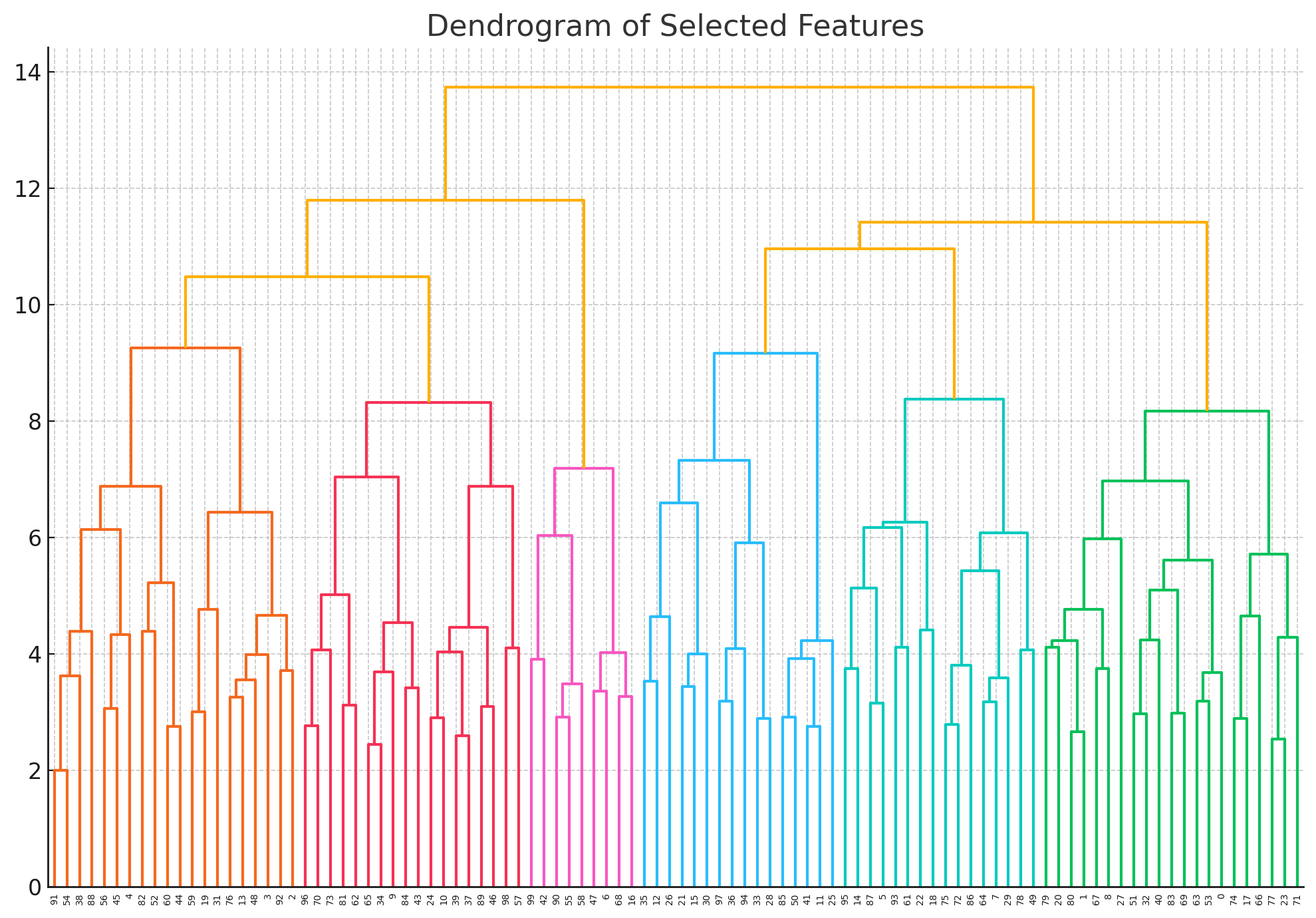}
\caption{Cluster similarity graph of features.}
\label{fig:data-3}
\end{figure}

\textbf{Similarity-Based Sampling:}  
For minority class samples $x_i$, new samples were generated by interpolating with $k$ nearest neighbors:
\begin{equation}
x_{new} = x_i + \lambda \cdot (x_{neighbor} - x_i)
\end{equation}
where $x_{neighbor}$ is a neighbor of $x_i$, and $\lambda \in [0, 1]$. This method maintained data similarity while improving learning. Figure \ref{fig:data-3} shows the cluster similarity graph.

\subsection{Outlier Detection and Handling}  
\begin{figure}[htbp]
\centering
\includegraphics[width=0.45\textwidth]{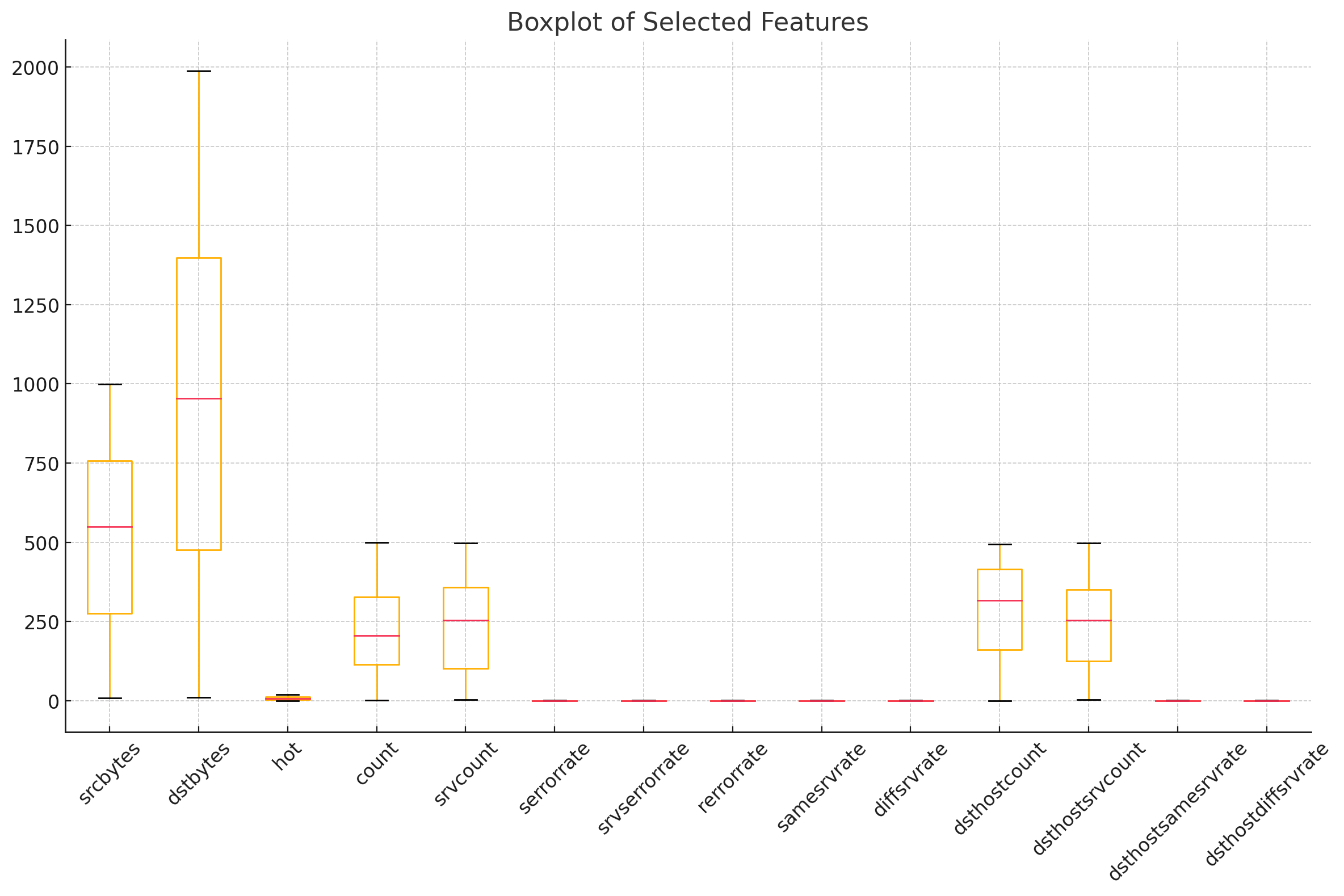}
\caption{Box plot of selected features.}
\label{fig:data-2}
\end{figure}

Outliers in the dataset can either represent attacks or noise. Figure \ref{fig:data-2} depicts outliers in certain features.

To differentiate between true anomalies and noise, two methods were used:
\begin{itemize}
    \item \textbf{Z-score Detection:} Outliers were flagged using the Z-score:
    \begin{equation}
    Z_i = \frac{x_i - \mu}{\sigma}
    \end{equation}
    Instances with $|Z_i| > 3$ were identified as outliers.
    
    \item \textbf{KNN-Based Detection:} KNN detected instances with large distances from their nearest neighbors as potential anomalies.
\end{itemize}
Outliers were either removed or labeled to retain meaningful data for training.

\subsection{Privacy Considerations}  
Network traffic data may include PII or confidential information, raising privacy concerns. We addressed this by anonymizing sensitive elements like IP addresses and credentials, ensuring regulatory compliance.

Additionally, differential privacy techniques were applied by adding controlled noise to the dataset to prevent re-identification, balancing privacy and data utility.

\section{Methodology}  
We propose a multi-model framework that integrates machine learning (KNN, SVM, XGBoost) and deep learning (ANN) to detect network anomalies and attacks. This framework accommodates both binary and multi-class classification, aiming to improve performance through preprocessing, advanced sampling techniques, and feature importance analysis to effectively manage small and imbalanced datasets. The model's performance is evaluated using accuracy, precision, recall, and F1-score, utilizing cross-entropy and focal loss functions. Additionally, privacy-preserving measures are incorporated to ensure data security.

\begin{figure}[htbp]
\centering
\includegraphics[width=0.5\textwidth]{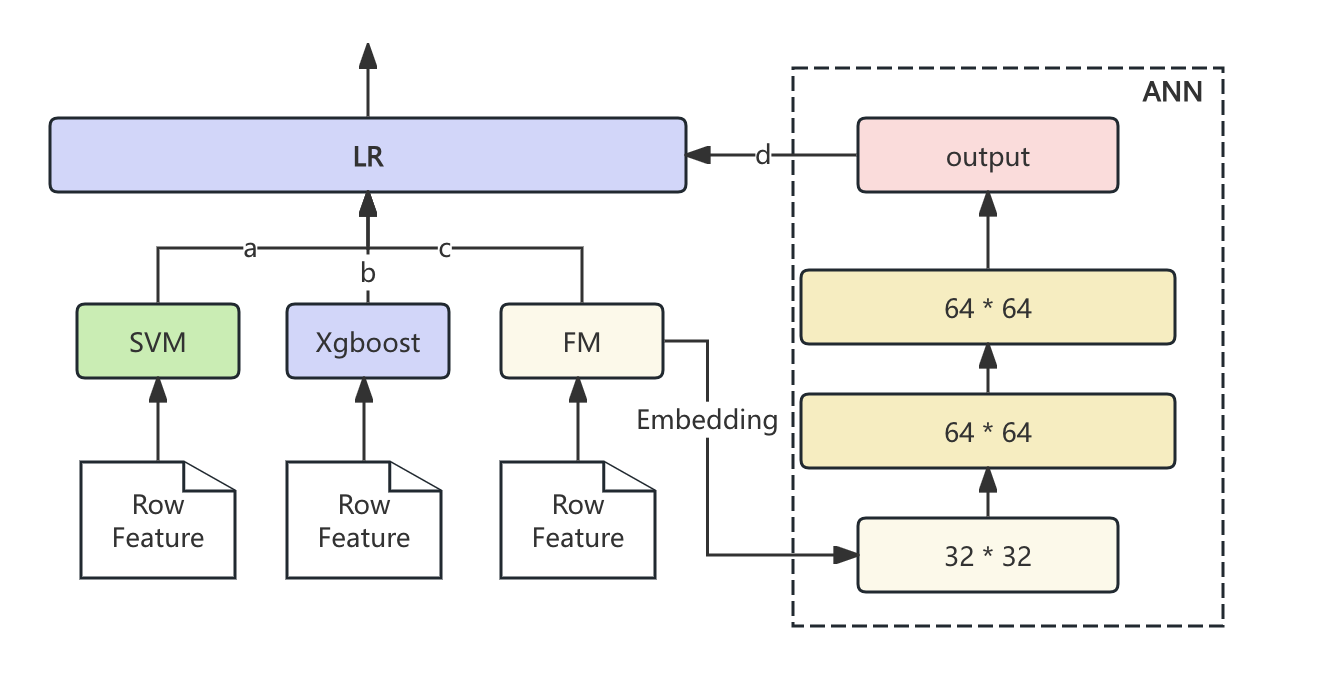}
\caption{Overall model process.}
\label{fig:model}
\end{figure}

The entire process is illustrated in Figure \ref{fig:model}.

\subsection{Privacy-Preserving Machine Learning Techniques}  
To address privacy concerns in models using sensitive network data, we applied several techniques.

\textbf{Federated Learning:}  
Models were trained across decentralized devices, sharing only updates instead of raw data.

\textbf{Secure Multi-Party Computation (SMPC):}  
SMPC enabled collaborative computations while keeping inputs private.

\textbf{Differential Privacy:}  
Noise was added to gradients or weights during training to protect individual data points, ensuring privacy.

\subsection{K-Nearest Neighbors (KNN)}  
\begin{figure}[htbp]
\centering
\includegraphics[width=0.4\textwidth]{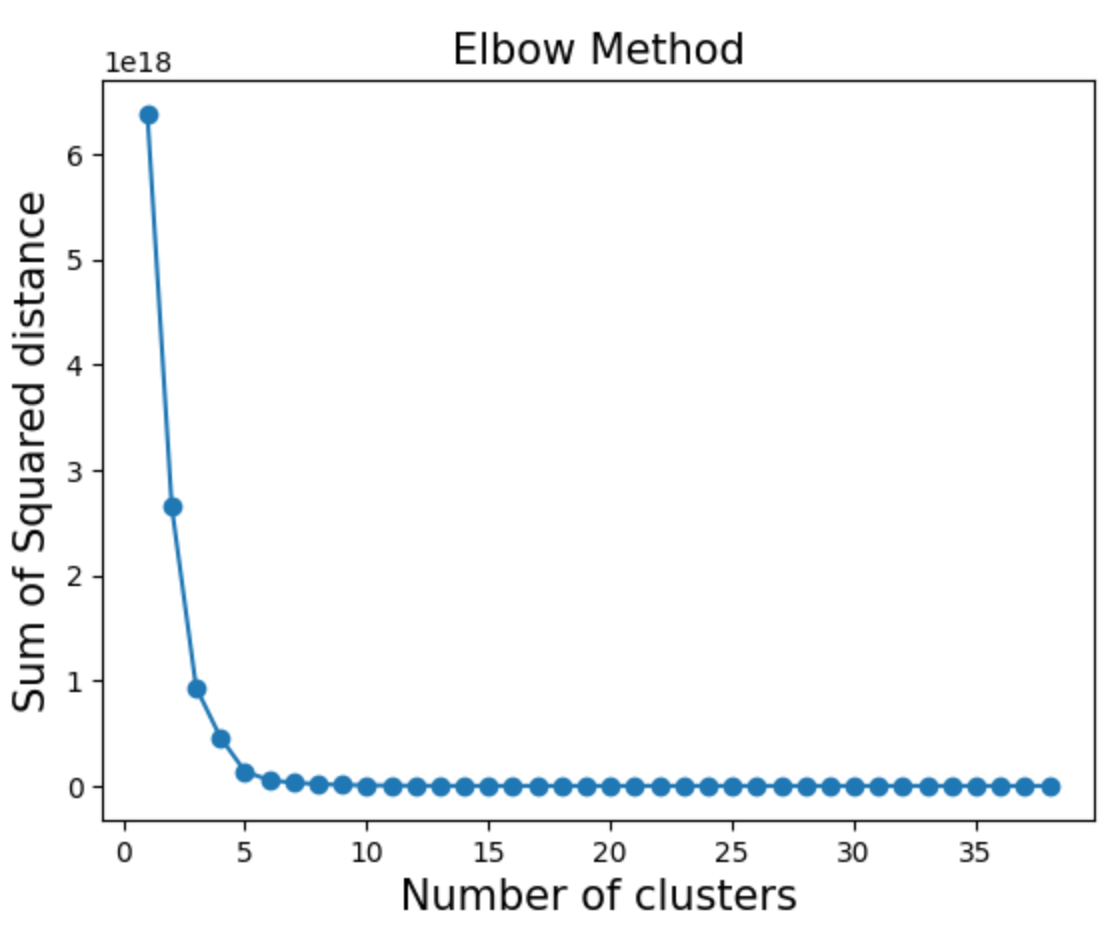}
\caption{KNN trained in 5-class with $k=5$.}
\label{fig:knn}
\end{figure}

We used KNN for clustering and feature importance, with $k=5$ for the 5-class classification. The Euclidean distance between feature vectors $x$ and $y$ is:
\begin{equation}
d(x, y) = \sqrt{\sum_{i=1}^{m} (x_i - y_i)^2}
\end{equation}
where $m$ is the number of features. KNN predicts labels by majority voting among the $k$ nearest neighbors:
\begin{equation}
\hat{y} = \arg\max_{c \in C} \sum_{i=1}^{k} \mathbb{I}(y_i = c)
\end{equation}
where $C$ is the set of classes. Figure \ref{fig:knn} shows KNN with $k=5$.

\subsection{Factorization Machines (FM)}  
FM utilizes embedding vectors from KNN results to capture feature interactions. The predicted output is:
\begin{equation}
\hat{y} = w_0 + \sum_{i=1}^{m} w_i x_i + \sum_{i=1}^{m} \sum_{j=i+1}^{m} \langle \mathbf{v}_i, \mathbf{v}_j \rangle x_i x_j
\end{equation}
where $w_0$ is the bias term, $w_i$ are first-order weights, and $\mathbf{v}_i, \mathbf{v}_j$ are embedding vectors with dot product $\langle \mathbf{v}_i, \mathbf{v}_j \rangle$.

\subsection{Artificial Neural Networks (ANN)}  
ANN learns from FM's embedding vectors using an architecture with input, hidden, and output layers. The input embedding vector $\mathbf{z}$ is:
\begin{equation}
\mathbf{z} = \mathbf{W}_e \mathbf{X} + \mathbf{b}_e
\end{equation}
where $\mathbf{W}_e$, $\mathbf{X}$, and $\mathbf{b}_e$ denote the embedding matrix, input matrix, and bias.

Each hidden layer applies a non-linear activation:
\begin{equation}
\mathbf{h}^{(l)} = f(\mathbf{W}^{(l)} \mathbf{h}^{(l-1)} + \mathbf{b}^{(l)})
\end{equation}
with weights $\mathbf{W}^{(l)}$, biases $\mathbf{b}^{(l)}$, and activation function $f(\cdot)$ (typically ReLU). The output layer uses softmax for multi-class classification:
\begin{equation}
\hat{y}_i = \frac{\exp(\mathbf{h}_i)}{\sum_{j=1}^{C} \exp(\mathbf{h}_j)}
\end{equation}
where $C$ is the number of classes.

\subsection{Support Vector Machine (SVM)}  
SVM separates classes by solving the optimization problem:
\begin{equation}
\min_{\mathbf{w}, b} \frac{1}{2} \|\mathbf{w}\|^2 + C \sum_{i=1}^{n} \max(0, 1 - y_i (\mathbf{w} \cdot \mathbf{x}_i + b))
\end{equation}
where $\mathbf{w}$ is the weight vector, $b$ is the bias, $C$ is the regularization parameter, $y_i$ is the label, and $\mathbf{x}_i$ is the input. SVM aims to maximize the class separation margin.

\subsection{XGBoost}  
XGBoost enhances classification by combining weak learners through gradient boosting. The objective function includes loss and regularization terms:
\begin{equation}
\mathcal{L}(\theta) = \sum_{i=1}^{n} \ell(\hat{y}_i, y_i) + \sum_{k=1}^{K} \Omega(f_k)
\end{equation}
where $\ell(\hat{y}_i, y_i)$ is the loss (e.g., cross-entropy), and $\Omega(f_k)$ regularizes each tree $f_k$. The model iteratively adds trees to minimize the objective.

\subsection{Ensemble Model}  
The ensemble prediction combines KNN, ANN, SVM, and XGBoost outputs using Logistic Regression:
\begin{equation}
\hat{y} = \sigma(\mathbf{w}^T \mathbf{o} + b)
\end{equation}
where $\mathbf{o}$ is the vector of model outputs, $\mathbf{w}$ is the weight vector, $b$ is the bias, and $\sigma(\cdot)$ is the sigmoid for binary classification or softmax for multi-class classification.

\subsection{Security of the Model}  
Securing the machine learning model is essential to prevent adversarial attacks.

\textbf{Adversarial Training:}  
Robustness was improved by training with adversarial examples.

\textbf{Model Hardening:}  
Regularization and pruning reduced model complexity, enhancing resilience.

\textbf{Secure Model Deployment:}  
The model was deployed using secure protocols, encrypted parameters, and access controls to ensure secure communication.

\subsection{Loss Functions}  
For classification, cross-entropy loss is used, defined for binary classification as:
\begin{equation}
\mathcal{L}_{CE} = -\left( y \log(\hat{y}) + (1 - y) \log(1 - \hat{y}) \right)
\end{equation}
where $y$ is the true label and $\hat{y}$ is the predicted probability.

To handle class imbalance, focal loss is applied for multi-class scenarios:
\begin{equation}
\mathcal{L}_{focal} = -\alpha (1 - \hat{y})^\gamma \log(\hat{y})
\end{equation}
where $\alpha$ scales the loss, and $\gamma$ emphasizes harder-to-classify samples.

\section{Evaluation Metrics}
Model performance is assessed using the following metrics:

\textbf{Accuracy:}
\begin{equation}
\text{Accuracy} = \frac{TP + TN}{TP + TN + FP + FN}
\end{equation}

\textbf{Precision:}
\begin{equation}
\text{Precision} = \frac{TP}{TP + FP}
\end{equation}

\textbf{Recall:}
\begin{equation}
\text{Recall} = \frac{TP}{TP + FN}
\end{equation}

\textbf{F1-score:}
\begin{equation}
\text{F1} = 2 \times \frac{\text{Precision} \times \text{Recall}}{\text{Precision} + \text{Recall}}
\end{equation}

\begin{figure}[htbp]
\centering
\includegraphics[width=0.5\textwidth]{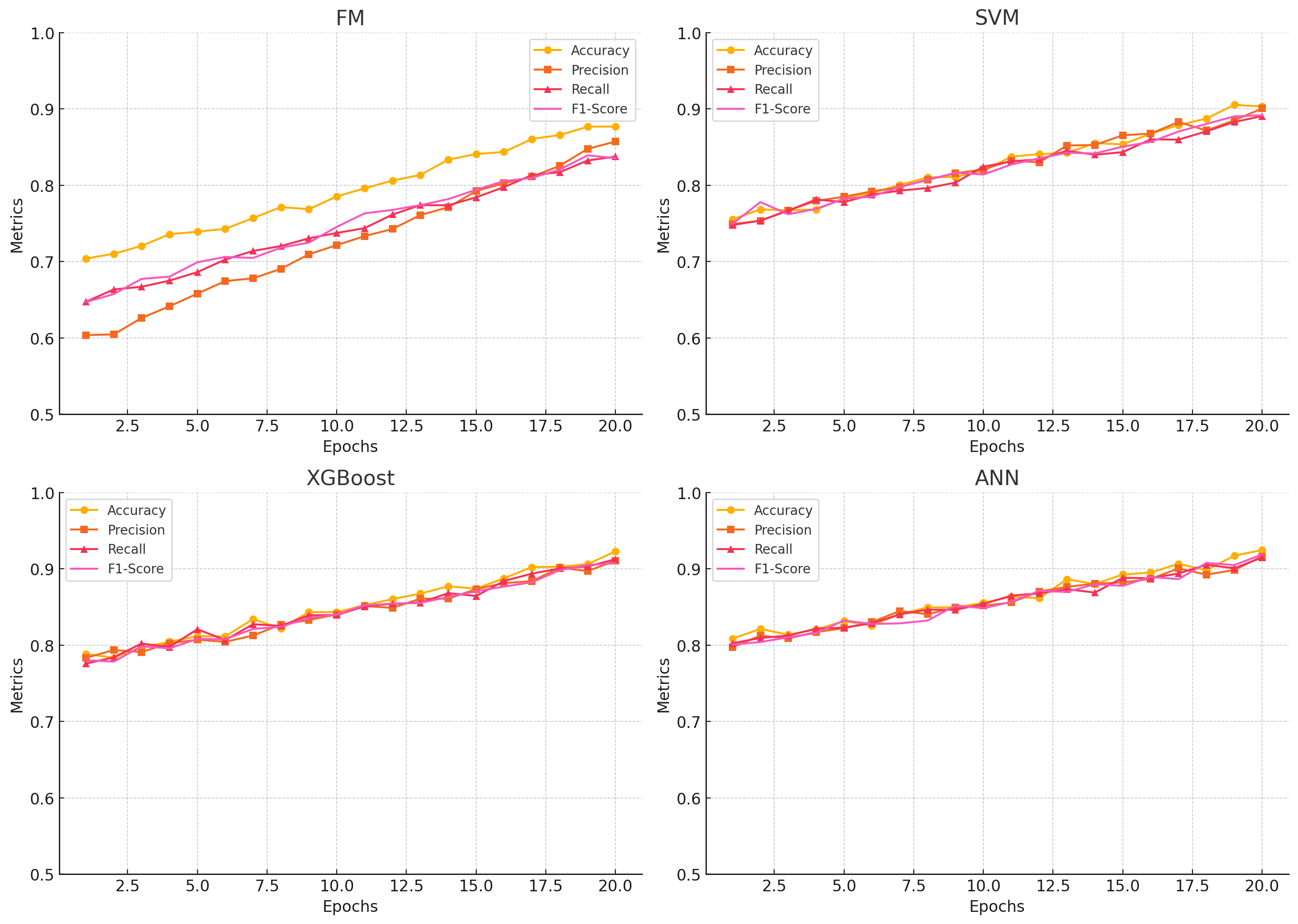}
\caption{Model indicator change chart.}
\label{fig:metric}
\end{figure}

Figure \ref{fig:metric} shows changes in model indicators.

Table~\ref{results_table} summarizes the performance comparison across models.
\begin{table}[ht]
\centering
\caption{Performance Comparison of Different Models}
\begin{tabular}{|c|c|c|c|c|}
\hline
\textbf{Model} & \textbf{Accuracy} & \textbf{Precision} & \textbf{Recall} & \textbf{F1-Score} \\
\hline
FM & 88.3\% & 85.4\% & 83.9\% & 84.6\% \\
\hline
SVM & 90.2\% & 89.7\% & 88.6\% & 89.1\% \\
\hline
XGBoost & 91.5\% & 90.6\% & 91.1\% & 90.8\% \\
\hline
ANN & 92.1\% & 91.7\% & 91.3\% & 91.5\% \\
\hline
Ensemble & \textbf{94.3\%} & \textbf{93.9\%} & \textbf{93.2\%} & \textbf{93.5\%} \\
\hline
\end{tabular}
\label{results_table}
\end{table}

\section{Conclusion}  
This paper proposed a multi-model approach for network intrusion detection, handling both binary and multi-class classification. KNN was used for feature analysis, ANN for embedding learning, and the outputs of SVM, XGBoost, and ANN were combined via Logistic Regression to create a robust ensemble model. Privacy-preserving techniques and security measures against adversarial attacks were also incorporated.
Experimental results demonstrated that the ensemble model outperformed individual models while ensuring privacy and security, contributing to improved network intrusion detection and user data protection.

 \bibliographystyle{IEEEtran}
    \bibliography{references}

\begin{thebibliography}{1}
\providecommand{\url}[1]{#1}
\csname url@samestyle\endcsname
\providecommand{\newblock}{\relax}
\providecommand{\bibinfo}[2]{#2}
\providecommand{\BIBentrySTDinterwordspacing}{\spaceskip=0pt\relax}
\providecommand{\BIBentryALTinterwordstretchfactor}{4}
\providecommand{\BIBentryALTinterwordspacing}{\spaceskip=\fontdimen2\font plus
\BIBentryALTinterwordstretchfactor\fontdimen3\font minus \fontdimen4\font\relax}
\providecommand{\BIBforeignlanguage}[2]{{%
\expandafter\ifx\csname l@#1\endcsname\relax
\typeout{** WARNING: IEEEtran.bst: No hyphenation pattern has been}%
\typeout{** loaded for the language `#1'. Using the pattern for}%
\typeout{** the default language instead.}%
\else
\language=\csname l@#1\endcsname
\fi
#2}}
\providecommand{\BIBdecl}{\relax}
\BIBdecl

\bibitem{yang2023achieving}
M.~Yang, S.~Liu, J.~Xu, G.~Tan, C.~Li, and L.~Song, ``Achieving privacy-preserving cross-silo anomaly detection using federated xgboost,'' \emph{Journal of the Franklin Institute}, vol. 360, no.~9, pp. 6194--6210, 2023.

\bibitem{maniriho2020anomaly}
P.~Maniriho, E.~Niyigaba, Z.~Bizimana, V.~Twiringiyimana, L.~J. Mahoro, and T.~Ahmad, ``Anomaly-based intrusion detection approach for iot networks using machine learning,'' in \emph{2020 international conference on computer engineering, network, and intelligent multimedia (CENIM)}.\hskip 1em plus 0.5em minus 0.4em\relax IEEE, 2020, pp. 303--308.

\bibitem{liu2020anomaly}
Z.~Liu, N.~Thapa, A.~Shaver, K.~Roy, X.~Yuan, and S.~Khorsandroo, ``Anomaly detection on iot network intrusion using machine learning,'' in \emph{2020 International conference on artificial intelligence, big data, computing and data communication systems (icABCD)}.\hskip 1em plus 0.5em minus 0.4em\relax IEEE, 2020, pp. 1--5.

\bibitem{nguyen2020poisoning}
T.~D. Nguyen, P.~Rieger, M.~Miettinen, and A.-R. Sadeghi, ``Poisoning attacks on federated learning-based iot intrusion detection system,'' in \emph{Proc. Workshop Decentralized IoT Syst. Secur.(DISS)}, vol.~79, 2020.

\bibitem{shen2024financial}
Y.~Shen and P.~K. Zhang, ``Financial sentiment analysis on news and reports using large language models and finbert,'' \emph{arXiv preprint arXiv:2410.01987}, 2024.

\end{thebibliography}

\end{document}